\documentclass[10pt,twocolumn,letterpaper]{article}

\usepackage{iccv}
\usepackage{times}
\usepackage{epsfig}
\usepackage{graphicx}
\usepackage{amsmath}
\usepackage{amssymb}

\usepackage{lipsum}
\usepackage{color}
\newcommand{\smalltitle}[1]{\vspace{0.2em}\noindent \textbf{{#1}}}

\newcommand{\cF}{\mathcal{F}}
\newcommand{\cP}{\mathcal{P}}

\newcommand{\W}{\mathbf{W}}
\newcommand{\w}{\mathbf{w}}
\newcommand{\bb}{\mathbf{b}}

\newcommand{\bS}{\mathbf{S}}
\newcommand{\bx}{\mathbf{x}}
\newcommand{\by}{\mathbf{y}}
\newcommand{\bz}{\mathbf{z}}

\newcommand{\Var}{\operatorname{Var}}
\newcommand{\E}{\operatorname{E}}
\newcommand{\Loss}{\mathcal{L}}

\newenvironment{myitemize}[1][]{
\begin{list}{{#1}} 
	{
		\setlength{\leftmargin}{1.2em}
		\setlength{\topsep}{0em}
		\setlength{\itemsep}{-0.2em}
}}
{\end{list}}

\usepackage[pagebackref=true,breaklinks=true,letterpaper=true,colorlinks,bookmarks=false]{hyperref}

\iccvfinalcopy 

\def\iccvPaperID{460} 

\ificcvfinal\fi
\begin{document}

\title{Learning Feature Pyramids for Human Pose Estimation}

\author{
Wei Yang$^{1}$ \quad 
Shuang Li$^{1}$ \quad 
Wanli Ouyang$^{1,2}$ \quad 
Hongsheng Li$^{1}$ \quad
Xiaogang Wang$^{1}$ \\
$^{1}$ Department of Electronic Engineering, The Chinese University of Hong Kong\\
$^{2}$~ School of Electrical and Information Engineering, The University of Sydney \\
{\tt\small \{wyang, sli, wlouyang, hsli, xgwang\}@ee.cuhk.edu.hk  } \\
}

\maketitle
\thispagestyle{empty}

\begin{abstract}
Articulated human pose estimation is a fundamental yet challenging task in computer vision. 
The difficulty is particularly pronounced in scale variations of human body parts when camera view changes or severe foreshortening happens. 
Although pyramid methods are widely used to handle scale changes at inference time, learning feature pyramids in deep convolutional neural networks (DCNNs) is still not well explored. 
In this work, we design a Pyramid Residual Module (PRMs) to enhance the invariance in scales of DCNNs. 
Given input features, the PRMs learn convolutional filters on various scales of input features, which are obtained with different subsampling ratios in a multi-branch network. 
Moreover, we observe that it is inappropriate to adopt existing methods to initialize the weights of multi-branch networks, 
which achieve superior performance than plain networks in many tasks recently. 
Therefore, we provide theoretic derivation to extend the current weight initialization scheme to multi-branch network structures. 
We investigate our method on two standard benchmarks for human pose estimation. 
Our approach obtains state-of-the-art results on both benchmarks. 
Code is available at~\url{https://github.com/bearpaw/PyraNet}.

\end{abstract}


\section{Introduction}

Localizing body parts for human body is a fundamental yet challenging task in computer vision, and it serves as an important basis for high-level vision tasks, \eg, activity recognition~\cite{yang2010recognizing,wang2013pose}, clothing parsing~\cite{yamaguchi2012parsing,yang2014clothing,liu2015matching}, human re-identification~\cite{zheng2017pose}, and human-computer interaction.
Achieving accurate localization, however, is difficult due to the highly articulated human body limbs, occlusion,  change of viewpoint, and foreshortening.

Significant progress on human pose estimation has been achieved by deep convolutional neural networks (DCNNs)~\cite{toshev2014deeppose,tompson2014joint,chen2014articulated,tompson2015efficient,pishchulin2016deepcut,wei2016convolutional,newell2016stacked}.  
In these methods, the DCNNs learn body part detectors from images warped to the similar scale based on human body size.  
At inference time, testing images should also be warped to the same scale as that for training images.

\begin{figure}[t]
	\begin{center}
		\includegraphics[width=1\linewidth]{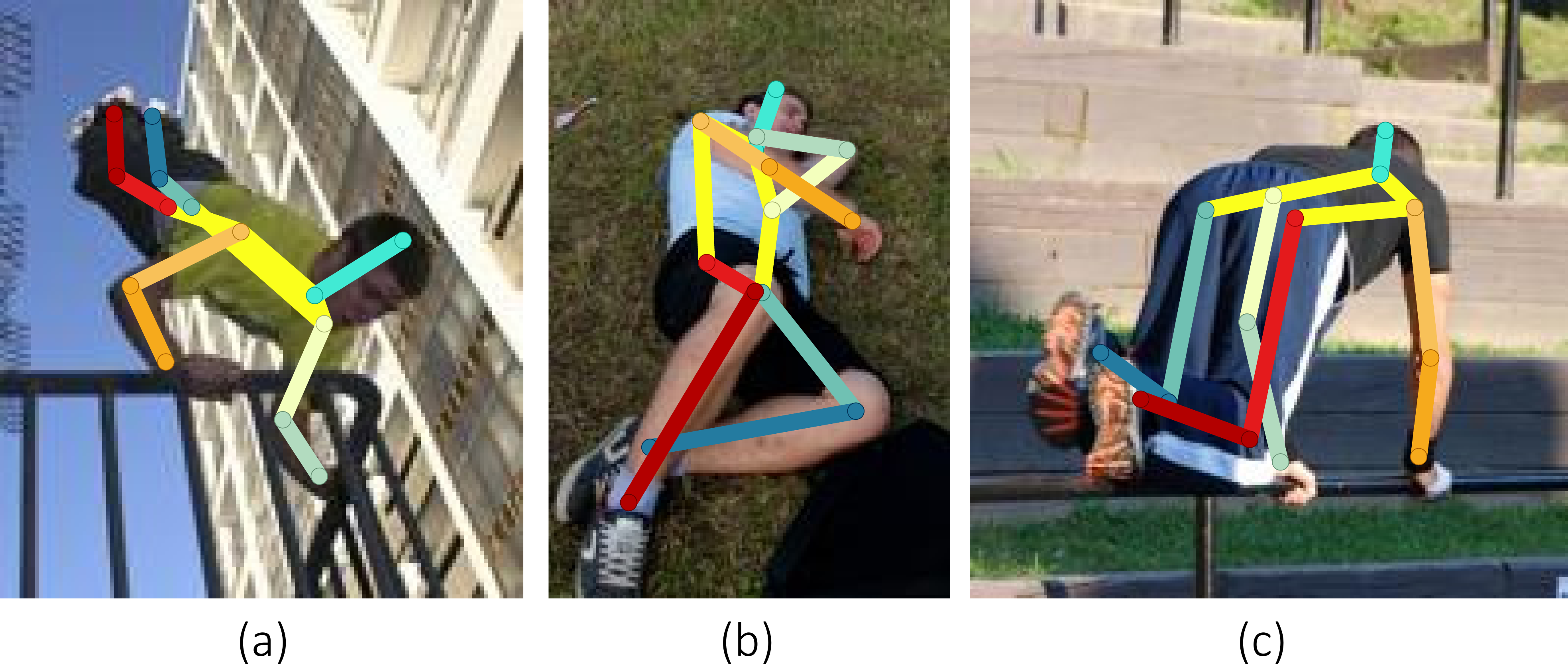}
	\end{center}
	\vspace{-1em}
	\caption{Our predictions on the LSP dataset~\cite{Johnson10}. When images are warped to approximately the same scale, scales of different body parts may still be  inconsistent due to camera view change and foreshortening. In (a), the scale of \textit{hand} and \textit{head} are larger than that of \textit{foot}. In (b), the scale of \textit{foot} is larger than that of \textit{head}. }
	\label{fig:motivation}
	\vspace{-1em}
\end{figure}

Although the right scale of the full human body is provided, scales for body parts may still be inconsistent due to inter-personal body shape variations and foreshortening caused by viewpoint change and body articulation. It results in difficulty for body part detectors to localize body parts. 
For example, severe foreshortening is present in Figure~\ref{fig:motivation}. 
When the images are warped to the same size according to human body scale, the \textit{hand} in Figure~\ref{fig:motivation} (a) has a larger scale than that in Figure~\ref{fig:motivation} (b). Therefore, the \textit{hand} detector that can detect the \textit{hand} in Figure~\ref{fig:motivation} (a) might not be able to detect the \textit{hand} in Figure~\ref{fig:motivation} (b) reliably. In DCNNs, this problem from scale change happens not only for high-level semantics in deeper layers, but also exists for low-level features in shallower layers.

%

To enhance the robustness of DCNNs against scale variations of visual patterns, we design a \textit{Pyramid Residual Module} to explicitly learn convolutional filters for building feature pyramids. 
Given input features, the Pyramid Residual Module obtains features of different scales via subsampling with different ratios. 
Then convolution is used to learn filters for features in different scales. 
The filtered features are upsampled to the same resolution and are summed together for the following processing. 
This Pyramid Residual Module can be used as building blocks in DCNNs for learning feature pyramids at different levels of the network.


There is a trend of designing networks with branches, \eg, Inception models~\cite{szegedy2015going,ioffe2015batch,szegedy2016rethinking,szegedy2016inception} and ResNets~\cite{he2016deep,he2016identity} for classification, ASPP-nets~\cite{chen2016deeplab} for semantic segmentation, convolutional pose machines~\cite{wei2016convolutional} and stacked hourglass networks~\cite{newell2016stacked} for human pose estimation, in which the input of a layer is from multiple other layers or the output of a layer is used by many other layers. Our pyramid residual module also has branches. We observe that the existing weight initialization scheme, \eg, MSR~\cite{he2015delving} and Xavier~\cite{glorot2010understanding} methods, are not proper for layers with branches. 
Therefore, we extend the current weight initialization scheme and provide theoretic derivation to show that the initialization of network parameters should take the number of branches into consideration.
We also show another issue in the residual unit~\cite{he2016identity}, where the variance of output of the residual unit accumulates as the depth increases. The problem is caused by the identity mapping. 

Since Hourglass network, also called conv-deconv structure, is an effective structure for pose estimation~\cite{newell2016stacked}, object detection~\cite{li2017zoom}, and pixel level tasks~\cite{chen2016single}, we use it as the basic structure in experiments. 
We observe a problem of using residual unit for Hourglass: when outputs of two residual units are summed up, the output variance is approximately doubled, which causes difficulty in optimization. 
We propose a simple but efficient way with negligible additional parameters to solve this problem.

The main contributions are three folds:
\begin{myitemize}
	\item[$\bullet$] We propose a \textit{Pyramid Residual Module}, which enhances the invariance in scales of deep models by learning feature pyramids in DCNNs with only a small increase of complexity.
	\item[$\bullet$] We identify the problem for initializing DCNNs including layers with multiple input or output branches. A weight initialization scheme is then provided, which can be used for many network structures including inception models~\cite{szegedy2015going,ioffe2015batch,szegedy2016rethinking,szegedy2016inception} and ResNets~\cite{he2016deep,he2016identity}.
	\item[$\bullet$] We observe that the problem of activation variance accumulation introduced by identity mapping may be harmful in some scenarios, \eg, adding outputs of multiple residual units implemented by identity mapping~\cite{he2016identity} together in the Hourglass structure. A simple yet effective solution is introduced for solving this issue. 
\end{myitemize}
We evaluate the proposed method on two popular human pose estimation benchmarks, and report state-of-the-art results. 
We also demonstrate the generalization ability of our approach on standard image classification task. 
Ablation study demonstrates the effectiveness of the pyramid residual module, the new initialization scheme, and the approach in handling drastic activation variance increase caused by adding residual units.

\section{Related Work}
\smalltitle{Human pose estimation. }
Graph structures, \eg, Pictorial structures~\cite{fischler1973representation,felzenszwalb2005pictorial,yang2011articulated} and loopy structures~\cite{ren2005recovering,tian2010fast,ferrari20092d}, have been broadly used to model the spatial relationships among body parts. 
All these methods were built on hand-crafted features such as HOG feature~\cite{dalal2005histograms}, and their performances relied heavily on image pyramid. 
Recently, deep models have achieved state-of-the-art results in human pose estimation~\cite{belagiannis2016recurrent,insafutdinov2016deepercut,bulat2016human,wei2016convolutional,newell2016stacked,chu2016structure,yang2016end,chu2017multi,cao2017realtime,papandreou2017towards}. 
Among them, DeepPose~\cite{toshev2014deeppose} is one of the first attempts on using DCNNs for human pose estimation. 								 
It regressed the coordinates of body parts directly, which suffered from the problem that image-to-locations is a difficult mapping to learn. 
Therefore, later methods modeled part locations as Gaussian peaks in score maps, and predicted the score maps with fully convolutional networks. 
In order to achieve higher accuracy, multi-scale testing on image pyramids was often utilized, which produced a multi-scale feature representation. 
Our method is a complementary to image pyramids.
																							 
On the other hand, to learn a model with strong scale invariance, a multi-branch network trained on three scales of image pyramid was proposed in~\cite{tompson2015efficient}. However, when image pyramids are used for training,  computation and memory linearly increases with the number of scales. In comparison, our pyramid residual module provides an efficient way of learning multi-scale features, with relatively small cost in computation and memory.

%

\smalltitle{DCNNs combining multiple layers. }
In contrast to traditional plain networks (\eg, AlexNet~\cite{krizhevsky2012imagenet} and VGG-nets~\cite{simonyan2014very}), multi-branch networks exhibit better performance on various vision tasks. 
In classification, the inception models~\cite{szegedy2015going,ioffe2015batch,szegedy2016rethinking,szegedy2016inception} are one of the most successful multi-branch networks. 
The input of each module is first mapped to low dimension by $1\times1$ convolutions, then transformed by a set of filters with different sizes to capture various context information and combined by concatenation. 
ResNet~\cite{he2016deep,he2016identity} can be regarded as a two-branch networks with one identity mapping branch. 
ResNeXt~\cite{xie2016aggregated} is an extension of ResNet, in which all branches share the same topology. The implicitly learned transforms are aggregated by summation. In our work, we use multi-branch network to explore another possibility: to learn multi-scale features. 

Recent methods in pose estimation, object detection and segmentation used features from multiple layers for making predictions~\cite{liu2016ssd,cai2016unified, hariharan2015hypercolumns,bell2016inside,newell2016stacked,chen2016deeplab}. Our approach is complementary to these works. 
For example, we adopt Hourglass as our basic structure, and replace its original residual units, which learn features from a single scale, with the proposed Pyramid Residual Module. 


\smalltitle{Weight initialization. }
Good initialization is essential for training deep models. 
Hinton and Salakhutdinov~\cite{hinton2006reducing} adopted the layer-by-layer pretraining strategy to train a deep autoencoder. 
Krizhevsky~\etal~\cite{krizhevsky2012imagenet} initialized the weight of each layer by drawing samples from a Gaussian distribution with zero mean and 0.01 standard deviation. 
However, it has difficulty in training very deep networks due to the instability of gradients~\cite{simonyan2014very}. 
Xavier initialization~\cite{glorot2010understanding} has provided a theoretically sound estimation of the variance of weight. 
It assumes that the weights are initialized close to zero, hence the nonlinear activations like Sigmoid and Tanh can be regarded as linear functions. 
This assumption does not hold for rectifier~\cite{nair2010rectified} activations. 
Thus He~\etal~\cite{he2015delving} proposed an initialization scheme for rectifier networks based on~\cite{glorot2010understanding}. 
All the above initialization methods, however, are derived for plain networks with only one branch. We identify the problem of the initialization methods when applied for multi-branch networks. 
An initialization scheme for networks with multiple branches is provided to handle this problem.

\section{Framework}

\begin{figure}
	\begin{center}
		\includegraphics[width=1\linewidth]{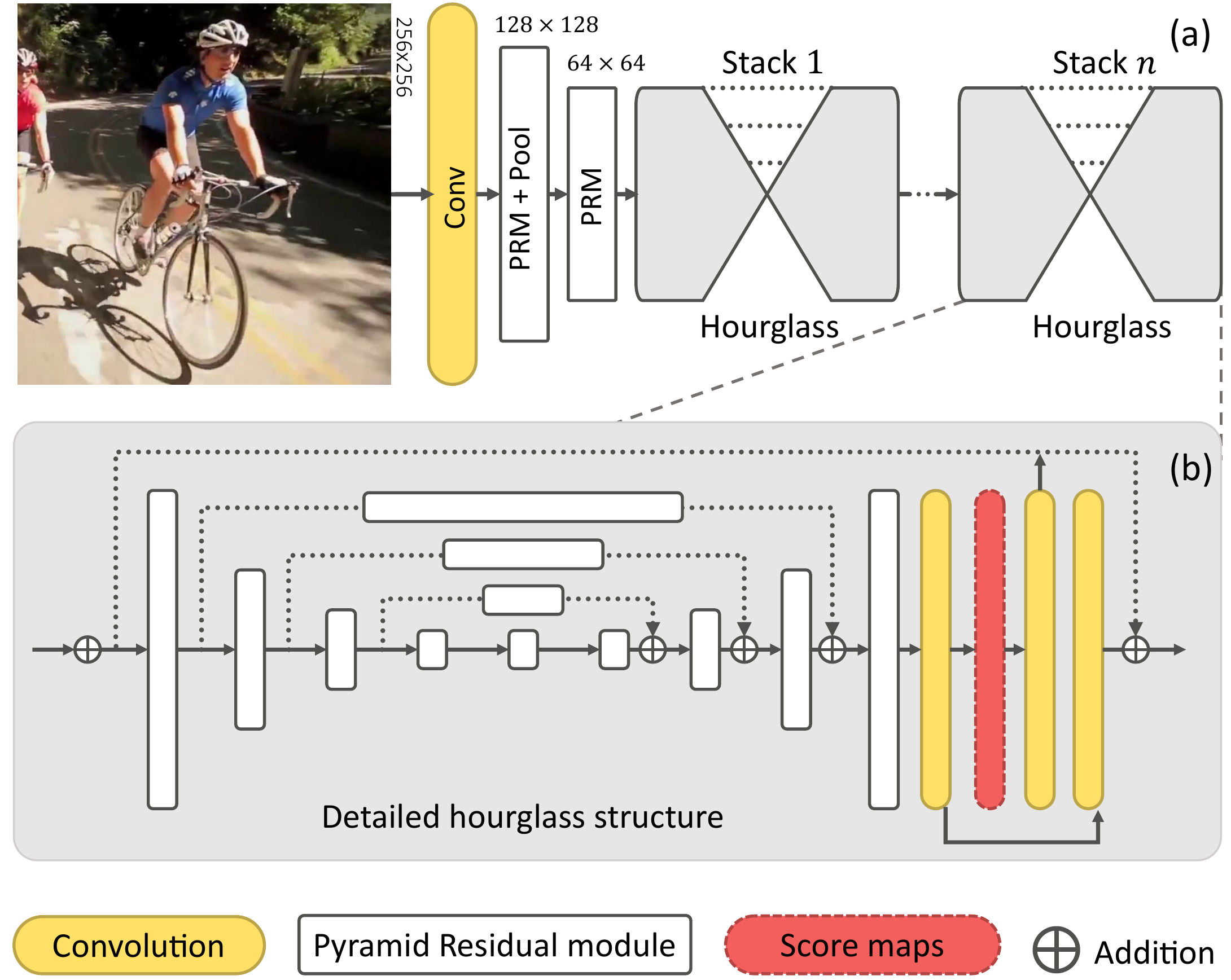}
	\end{center}
	\vspace{-0.5em}
	\caption{Overview of our framework. (a) demonstrates the network architecture, which has $n$ stacks of hourglass network. Details of each stack of hourglass is illustrated in (b). 
		Score maps of body joint locations are produced at the end of each hourglass, and a squared-error loss is also attached in each stack of hourglass.}
	\label{fig:framework}
	\vspace{-1em}
\end{figure}

An overview of the proposed framework is illustrated in Figure.~\ref{fig:framework}. 
We adopt the highly modularized stacked Hourglass Network~\cite{newell2016stacked} as the basic network structure to investigate feature pyramid learning for human pose estimation . 
The building block of our network is the proposed \textit{Pyramid Residual Module} (PRM). 
We first briefly review the structure of hourglass network. 
Then a detailed discussion of our pyramid residual module is presented.

\subsection{Revisiting Stacked Hourglass Network}
Hourglass network aims at capturing information at every scale in feed-forward fashion. It first performs bottom-up processing by subsampling the feature maps, and conducts top-down processing by upsampling the feature maps with the comination of higher resolution features from bottom layers, as demonstrated in Figure.~\ref{fig:framework}(b). 
This bottom-up, top-down processing is repeated for several times to build a ``stacked hourglass" network, with intermediate supervision at the end of each stack. 

In~\cite{newell2016stacked}, residual unit~\cite{he2016identity} is used as the building block of the hourglass network. 
However, it can only capture visual patterns or semantics at one scale. 
In this work, we use the proposed pyramid residual module as the building block for capturing multi-scale visual patterns or semantics.

\subsection{Pyramid Residual Modules (PRMs)}\label{sec:prm}

\begin{figure*}[t]
	\begin{center}
		\includegraphics[width=0.92\linewidth]{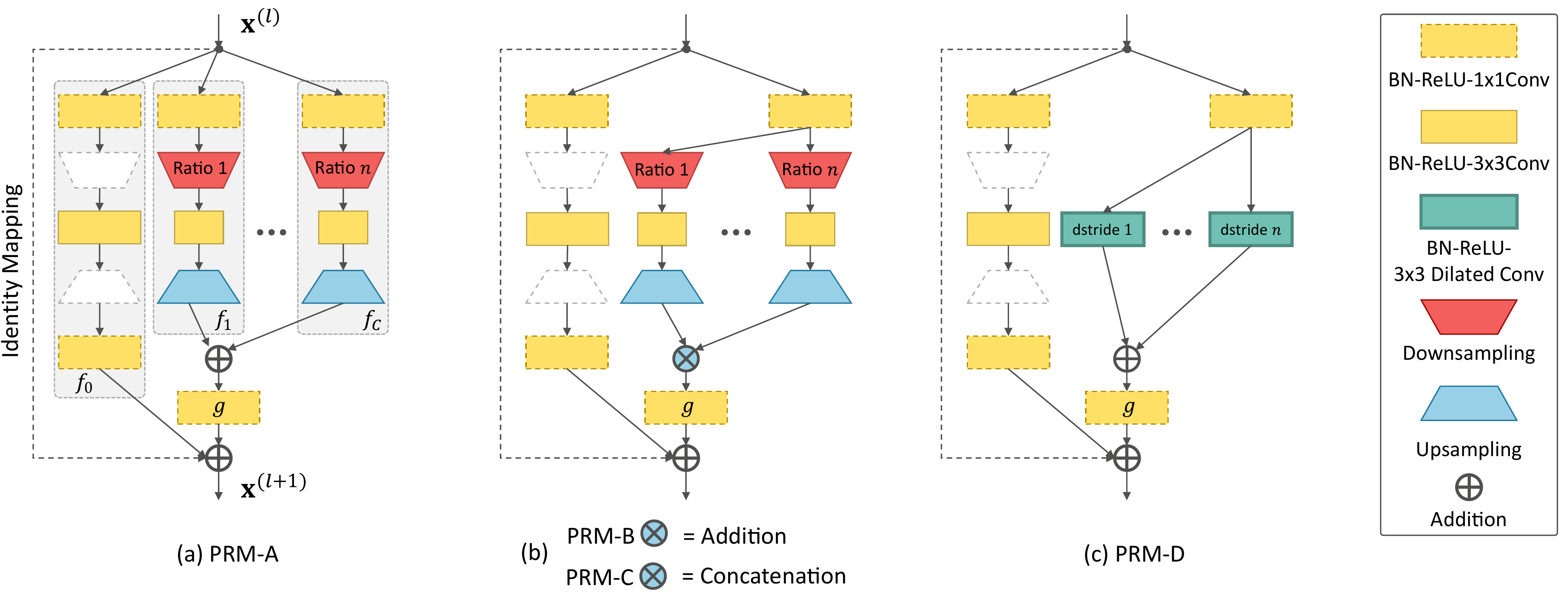}
	\end{center}
	\caption{Structures of PRMs. Dashed links indicate identity mapping. 
		(a) PRM-A produces separate input feature maps for different levels of pyramids, while (b) PRM-B uses shared input for all levels of pyramids. 
		PRM-C use concatenation instead of addition to combine features generated from pyramids, which is similar to inception models. 
		(c) PRM-D use dilated convolutions, which are also used in ASPP-net~\cite{chen2016deeplab}, instead of pooling to build the pyramid. 
		The dashed trapezoids mean that the subsampling and upsampling are skipped. }
	\label{fig:pyramidmodule}
	\vspace{-1em}
\end{figure*}

The objective is to learn feature pyramids across different levels of DCNNs. 
It allows the network to capture feature pyramids from primitive visual patterns to high-level semantics. 
Motivated by recent progress on residual learning~\cite{he2016deep,he2016identity}, we propose a novel Pyramid Residual Module (PRM), which is able to learn multi-scale feature pyramids.

The PRM explicitly learns filters for input features with different resolutions. 
Let $\bx^{(l)}$ and $\W^{(l)}$ be the input and the filter of the $l$-th layer, respectively. 
The PRM can be formulated as,
{\small
\begin{align}
	\bx^{(l+1)} = \bx^{(l)} + \cP(\bx^{(l)}; \W^{(l)}),
\end{align}
}
\!\!where $ \cP(\bx^{(l)}; \W^{(l)})$ is feature pyramids decomposed as:
{\small 
\begin{align}
	\cP(\bx^{(l)}; \W^{(l)}) = g\left(\sum_{c=1}^{C} f_c(\bx^{(l)}; \w^{(l)}_{f_c}); \w^{(l)}_g \right) +  f_0(\bx^{(l)}; \w^{(l)}_{f_0}).
\label{eq:prm}
\end{align}
}
\!\!The $C$ in (\ref{eq:prm}) denotes the number of pyramid levels, $f_c(\cdot)$ is the transformation for the $c$-th pyramid level, and $\W^{(l)} = \{\w^{(l)}_{f_c}, \w^{(l)}_g\}_{c=0}^C$ is the set of parameters. 
Outputs of transformations $f_c(\cdot)$ are summed up together, and further convolved by filters $g(\cdot)$. 
An illustration of the pyramid residual module is illustrated in Figure.~\ref{fig:pyramidmodule}. 
To reduce the computational and space complexity, each $f_c(\cdot)$ is designed as a bottleneck structure. 
For example, in Figure.~\ref{fig:pyramidmodule}, the feature dimension is reduced by a $1\times 1$ convolution, then new features are computed on a set of subsampled input features by $3\times3$ convolutions.
Finally, all the new features are upsampled to the same dimension and are summed together.

\smalltitle{Generation of input feature pyramids. }
Max-pooling or average-pooling are widely used in DCNNs to reduce the resolution of feature maps, and to encode the translation invariance. 
But pooling reduces the resolution too fast and coarse by a factor of an integer of at least two, which is unable to generate pyramids gently.   
In order to obtain input feature maps of different resolutions, we adopt the fractional max-pooling~\cite{graham2014fractional} to approximate the smoothing and subsampling process used in generating traditional image pyramids. 
The subsampling ratio of the $c$th level pyramid is computed as:
\begin{align}
	s_c = 2^{-M \frac{c}{C}}, \quad c = 0, \cdots, C, M \geq 1,
\end{align}
where $s_c \in [2^{-M}, 1]$ denotes the relative resolution compared with the input features. 
For example, when $c=0$, the output  has the same resolution as its input. 
When $ M=1, c=C$, the map has half resolution of its input. 
In experiments, we set $M=1$ and $C=4$, with which the lowest scale in pyramid is half the resolution of its input.

\subsection{Discussions} 

\smalltitle{PRM for general CNNs. } Our PRM is a general  module and can be used as the basic building block for various CNN architectures, \eg, stacked hourglass networks~\cite{newell2016stacked} for pose estimation, and Wide Residual Nets~\cite{zagoruyko2016WRN} and ResNeXt~\cite{xie2016aggregated} for image classification, as demonstrated in experiments.

\smalltitle{Variants in pyramid structure. } Besides using fractional max-pooling, convolution and upsampling to learn feature pyramids, as illustrated in Figure.~\ref{fig:pyramidmodule}(a-b), one can also use dilated convolution~\cite{chen2016deeplab,yu2016multi} to compute pyramids, as shown in Figure.~\ref{fig:pyramidmodule}(c)(PRM-D). 
The summation of features in pyramid can also replaced by concatenation, as shown in Figure.~\ref{fig:pyramidmodule}(b)(PRM-C). 
We discuss the performance of these variants in experiments, and show that the design in Figure.~\ref{fig:pyramidmodule}(b)(PRM-B) has comparable performance with others, while maintains relatively fewer parameters and smaller computational complexity.  

\smalltitle{Weight sharing. }
To generate the feature pyramids, traditional methods usually apply a same handcrafted filter, \eg, HOG, on different levels of image pyramids~\cite{adelson1984pyramid,felzenszwalb2010object}. 
This process corresponds to sharing the weights $\W^{(l)}_{f_c}$ across different levels of pyramid $f_c(\cdot)$, which is able to greatly reduce the number of parameters.

\smalltitle{Complexity. }
The residual unit used in~\cite{newell2016stacked} has 256-d input and output, which are reduced to 128-d within the residual unit. 
We adopt this structure for the branch with original scale (\ie, $f_0$ in Eq.(\ref{eq:prm})). 
Since features with smaller resolution contain relatively fewer information, we use fewer feature channels for branches with smaller scales. 
For example, given a PRM with five branches and 28 feature channels for branches with smaller scale (\ie, $f_1$ to $f_4$ in Eq.(\ref{eq:prm})), the increased complexity is about only $10\%$ compared with residual unit in terms of both parameters and GFLOPs.



\section{Training and Inference}
We use score maps to represent the body joint locations. Denote the ground-truth locations by $ \mathbf{z} = \{\bz_k \}_{k=1}^K$, where $\bz_k=(x_k, y_k)$ denotes the location of the $k$th body joint in the image. Then the ground-truth score map $\bS_k$ is generated from a Gaussian with mean $\bz_k$ and variance $\boldsymbol\Sigma$ as follows,
{\small
\begin{align}
\mathbf{S}_k(\mathbf{p}) \sim \mathcal{N}(\mathbf{z}_k, \boldsymbol\Sigma),
\end{align}
}
\!\!where $\mathbf{p}\in R^2$ denotes the location, and $\boldsymbol\Sigma$ is empirically set as an identity matrix $\textbf{I}$. Each stack of hourglass network predicts $K$ score maps, \ie $\mathbf{\hat{S}} = \{\mathbf{\hat{S}}_k\}_{k=1}^K$, for $K$ body joints. A loss is attached at the end of each stack defined by the squared error
{\small
\begin{align}
\mathcal{L} = \frac{1}{2} \sum_{n=1}^N \sum_{k=1}^{K} \|\mathbf{S}_k - \mathbf{\hat{S}}_k\|^2,
\end{align}
}
\!\!where $N$ is the number of samples.

During inference, we obtain the predicted body joint locations $\hat{\bz}_k$ from the predicted score maps generated from the last stack of hourglass by taking the locations with the maximum score as follows:
{\small
\begin{align}
\hat{\bz}_k = \arg \max_{\mathbf{p}} \mathbf{\hat{S}}_k(\mathbf{p}), \quad k = 1, \cdots, K.
\end{align}
}

\vspace{-1em}
\subsection{Initialization Multi-Branch Networks}


Initialization is essential to train very deep networks~\cite{glorot2010understanding,simonyan2014very,he2015delving}, especially for tasks of dense prediction, where Batch Normalization~\cite{ioffe2015batch} is less effective because of the small minibatch due to the large memory consumption of fully convolutional networks.  
Existing weight initialization methods~\cite{krizhevsky2012imagenet,glorot2010understanding,he2015delving} are designed upon the assumption of a plain networks without branches. 
The proposed PRM has multiple branches, and does not meet the assumption. 
Recent developed architectures with multiple branches, \eg, Inception models~\cite{szegedy2015going,ioffe2015batch,szegedy2016rethinking,szegedy2016inception} and ResNets~\cite{he2016deep,he2016identity}, are not plain network either. 
Hence we discuss how to derive a proper initialization for networks adding multiple branches. 
Our derivation mainly follows~\cite{glorot2010understanding,he2015delving}. 

\begin{figure}[t]
	\begin{center}
		\includegraphics[width=0.8\linewidth]{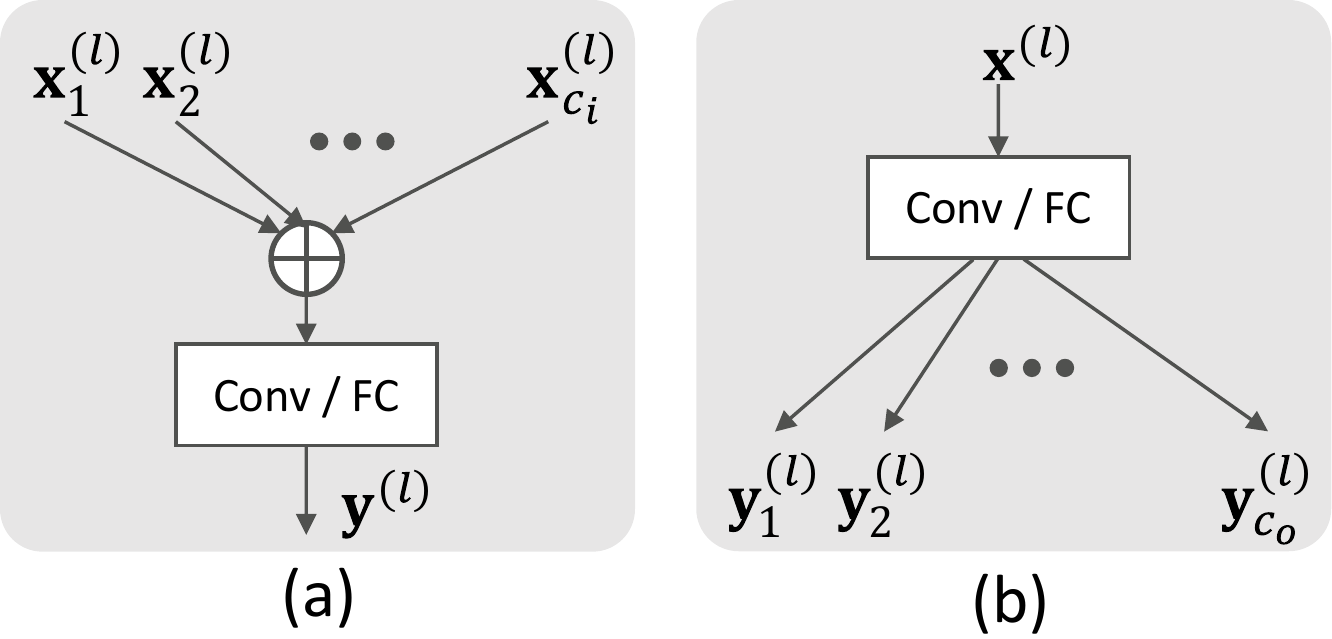}
	\end{center}
	\caption{Examples of multi-branch networks when (a) the inputs might be an addition of multiple branches, or (b) the output might be forwarded to multiple branches. 
	}
	\label{fig:indegree_outdegree}
	\vspace{-1em}
\end{figure}

\smalltitle{Forward propagation. }
Generally, multi-branch networks can be characterized by the number of input and output branches. 
Figure.~\ref{fig:indegree_outdegree} (a) shows an example where the $l$th layer has $C_i^{(l)}$ input branches and one output branch. Figure.~\ref{fig:indegree_outdegree} (b) shows an example where the  $l$th layer has one input branch and $C_o^{(l)}$ output branches. 
During forward propagation,  $C_{i}^{(l)}$ affects the variance for the output of the $l$th layer while $C_{o}^{(l)}$ does not. 
At the $l$th layer, assume there are $C_{i}^{(l)}$ input branches and $C_{o}^{(l)}$ output branches. There are $C_{i}^{(l)}$ input vectors $\{\bx_c^{(l)}|c=1, \ldots, C_i^{(l)}\}$.
Take fully-connected layer for example, a response is computed as:
{\small
\begin{align}
\by^{(l)}   &= \W^{(l)}\sum_{c=1}^{C^{(l)}_i}\bx^{(l)}_c + \bb^{(l)}, \\
\bx^{(l+1)} &= f\left(\by^{(l)} \right),\label{eq:relu}
\end{align}
}
\!\!where $f(\cdot)$ is the non-linear activation function.

As in~\cite{glorot2010understanding,he2015delving}, we assume that $\W^{(l)}$ and $\bx^{(l)}$ are both independent and identically distributed (i.i.d.), and they are independent of each other. Therefore, we respectively denote $y^{(l)}, x^{(l)}$ and $w^{(l)}$ as  the element in $\by^{(l)}, \bx^{(l)}$ and $\W^{(l)}$.  Then we have,
{\small
\begin{align}
\Var \left[y^{(l)}\right] = C^{(l)}_in^{(l)}_i \Var\left[w^{(l)} x^{(l)}\right] ,
\end{align}
}
\!\!where $n_i^{(l)}$ is the number of elements in $\bx_c^{(l)}$ for $c=1, \ldots, C_i^{(l)}$. Suppose $w^{(l)}$ has zero mean. The variance for the product of independent variables above is as follows:
{\small
\begin{align*}
\Var \left[y^{(l)}\right] = C^{(l)}_in^{(l)}_i \Var\left[w^{(l)}\right] \E\left[\left(x^{(l)}\right)^2\right] \\
= \alpha C^{(l)}_in^{(l)}_i\Var\left[w^{(l)}\right] \Var \left[y^{(l-1)}\right],
\end{align*}
}
\!\!where $\alpha$ depends on the activation function $f$ in (\ref{eq:relu}). $\alpha=0.5$ for ReLU and $\alpha=1$ for Tanh and Sigmoid. 
In order to make the variances of the output $y^{(l)}$ approximately the same for different layers $l$, the following condition should be satisfied:
{\small
\begin{align}
\alpha C^{(l)}_in^{(l)}_i\Var\left[w^{(l)}\right] = 1.
\end{align}
}
\!\!Hence in initialization, a proper variance for $W^{(l)}$ should be $1/(\alpha C^{(l)}_in^{(l)}_i)$.

\smalltitle{Backward propagation. }
Denote $\frac{\partial \Loss}{\partial \bx^{(l)}}$ and $ \frac{\partial \Loss}{\partial \by^{(l)}}$ by $\Delta \bx^{(l)}$ and $\Delta \by^{(l)}$ respectively. During backward propagation, the gradient is computed by chain rule,
{\small
\begin{align}
\Delta \bx^{(l)} &= \sum_{c=1}^{C^{(l)}_o}\W^{(l)T} \Delta \by^{(l)}, \\
\Delta \by^{(l)} &= f'(\by^{(l)})\Delta \bx^{(l+1)}. \label{eq：delta_y}
\end{align}
}
\!\!Suppose $w^{(l)}$ and $\Delta y^{(l)}$ are i.i.d. and independent of each other, then $\Delta x^{(l)}$ has zero mean when $w^{(l)}$ is initialized with zero mean and symmetric with small magnitude. 
Let $n_o^{(l)}$ denote the number of output neurons. Then we have,
{\small
\begin{align}
\Var\left[\Delta x^{(l)}\right] = C_o^{(l)}n_o^{(l)}\Var[w^{(l)}]\Var[\Delta y^{(l)}]. \label{eq:var_delta_x1}
\end{align}
}
\!\!Denote $\E(f'(y^{(l)})) = \alpha$.  $\alpha=0.5$ for ReLU and $\alpha=1$ for Tanh and Sigmoid. 
We further assume that $f'(y^{(l)})$ and $\Delta x^{(l)}$ are independent of each other, then from Eq. (\ref{eq：delta_y}), we have $\E\left[\Delta \by^{(l)}\right] =\alpha \E\left[\Delta \bx^{(l+1)}\right].$
Then we can derive that $\Var[\Delta y^{(l)}] = \E[(\Delta y^{(l)})^2] = \alpha\Var[x^{(l+1)}]$.  
Therefore, from Eq.(\ref{eq:var_delta_x1}) we have,
{\small
\begin{align}
\Var\left[\Delta x^{(l)}\right] = \alpha C^{(l)}_on^{(l)}_o\Var[w^{(l)}]\Var[\Delta x^{(l+1)}]. \label{eq:var_delta_x2}.
\end{align} 
}
\!\!To ensure $\Var[\Delta x^{(l)}] = \Var[\Delta x^{(l+1)}]$, we must have $\Var[w^{(l)}] = 1/(\alpha C^{(l)}_on^{(l)}_o)$.

In many cases, $C^{(l)}_in^{(l)}_i \neq C^{(l)}_o n^{(l)}_o$. As in~\cite{glorot2010understanding}, a compromise between the forward and backward constraints is to have,
{\small
\begin{align}
\Var[w^{(l)}] = \frac{1}{\alpha^2 (C^{(l)}_i n^{(l)}_i + C^{(l)}_on^{(l)}_o)}, \quad \forall l.
\label{eq:iniSol}
\end{align}
}

\smalltitle{Special case.} For plain networks with one input and one output branch, we have $C^{(l)}_i=C^{(l)}_o=1$ in (\ref{eq:iniSol}). In this case, the result in (\ref{eq:iniSol}) degenerates to the conclusions obtained for Tanh and Sigmoid in~\cite{glorot2010understanding}  and the conclusion in~\cite{he2015delving} for ReLU.

\smalltitle{General case.} In general, a network with branches would have $C^{(l)}_i\neq 1$ or $C^{(l)}_o\neq 1$ for some $l$s. Therefore, the number of input branches and output branches should be taken into consideration when initializing parameters. 
Specifically, if several multi-branch layers are stacked together without other operations (\eg, batch normalization,convolution, ReLU, \etc.), the output variance would be increased approximately $\prod_{l} C^{(l)}_i$ times by using Xavier~\cite{glorot2010understanding} or MSR~\cite{he2015delving} initialization.

\subsection{Output Variance Accumulation}\label{sec:control_va}
%
%
\begin{figure}[t]
	\begin{center}
		\includegraphics[width=0.8\linewidth]{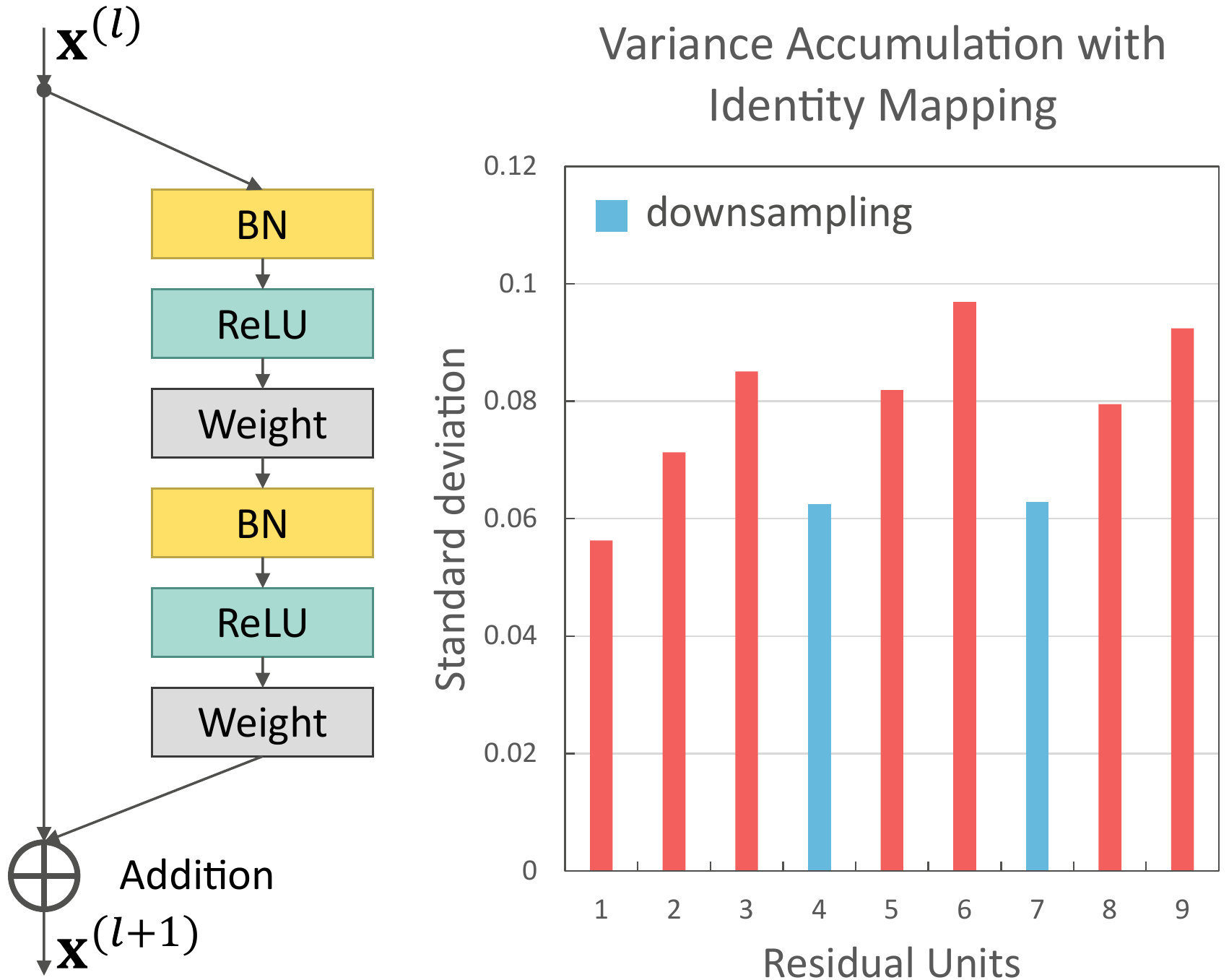}
	\end{center}
	\caption{ Response variances accumulate in ResNets. This accumulation can be reset (blue bar) when the identity mappings are replaced by convolution or batch normalization (\ie, when the feature channels of feature resolutions changes between input and output features).}
	\label{fig:variance_accumulation}
	\vspace{-1.5em}
\end{figure}
Residual learning~\cite{he2016deep,he2016identity} allows us to train extremely deep neural networks due to  identity mappings. 
But it is also the source of its drawbacks: identity mapping keeps increasing the variances of responses when the network goes deeper, which increases the difficulty of optimization. 

The response of the residual unit is computed as follows:
{\small
\begin{align}
\bx^{(l+1)} = \bx^{(l)} + \cF\left(\bx^{(l)}; \W^{(l)}\right),
\end{align}
}
\!\!where $\cF$ denotes the residual function, \eg, a bottleneck structure with three convolutions ($1\times1\rightarrow 3\times 3 \rightarrow 1\times 1$). 
Assume $\bx^{(l)}$ and $\cF\left(\bx^{(l)}; \W^{(l)}\right)$ are uncorrelated, then the variance of the response of residual unit is as
{\small
\begin{align}
\Var\left[\bx^{(l+1)}\right] &= \Var\left[\bx^{(l)}\right] + \Var\left[\cF\left(\bx^{(l+1)}; \W^{(l)}\right)\right] \nonumber \\
&> \Var\left[\bx^{(l)}\right],
\end{align}
}
\!\!where $\Var\left[\cF\left(\bx^{(l+1)}; \W^{(l)}\right)\right]$ is positive. 

\begin{figure}[t]
	\begin{center}
		\includegraphics[width=0.8\linewidth]{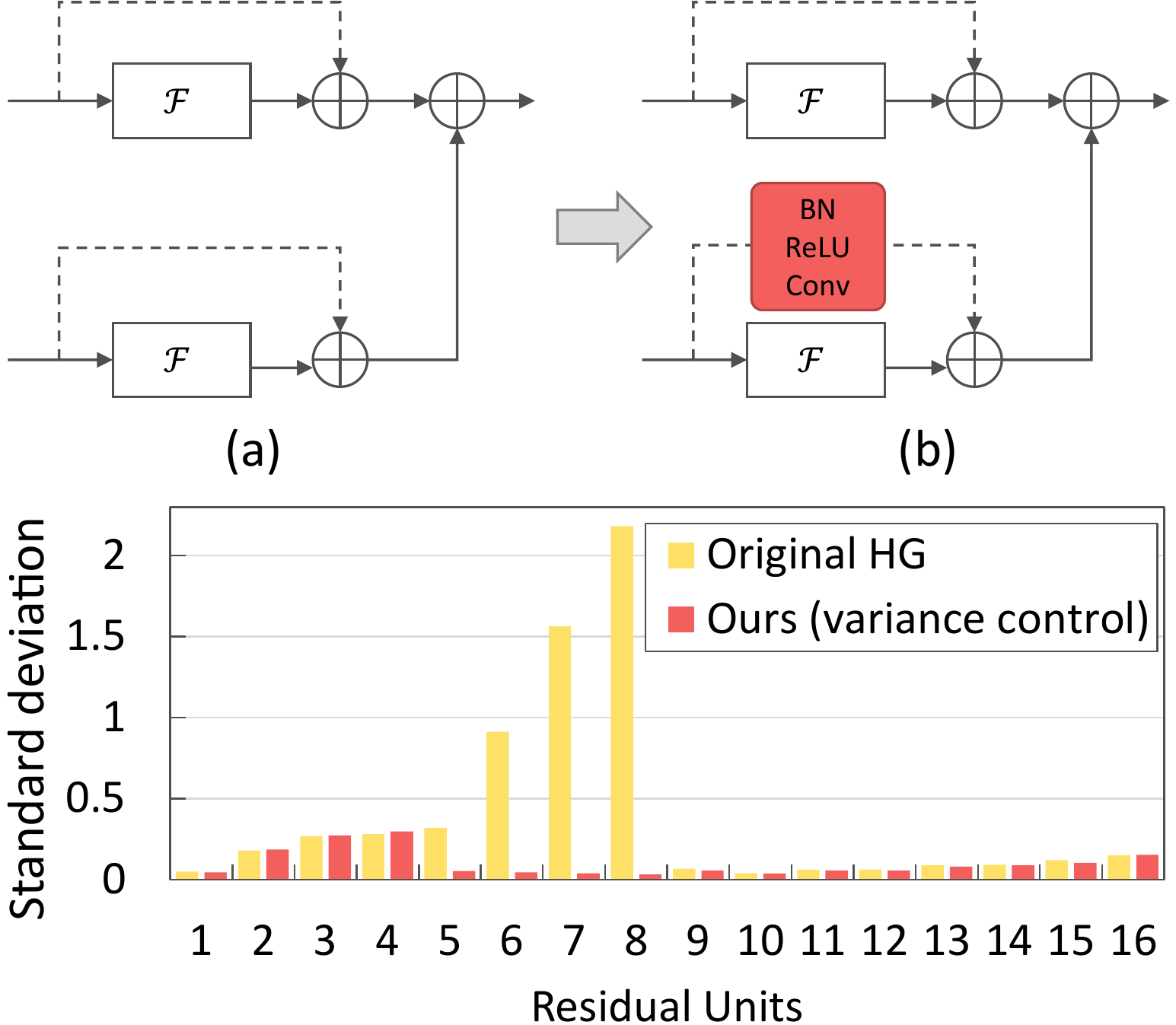}
	\end{center}
		\vspace{-0.3em}
	\caption{ 
	Top: (a) Addition of outputs of two identity mappings. 
	(b) One identity mapping is replaced by a BN-ReLU-Conv block. 
	Bottom: Statistics of response variances of the original hourglass network (yellow bar) and our structure (b) (red bar).
	}
	\label{fig:Useconv}
	\vspace{-1em}
\end{figure}

In ~\cite{he2016deep,he2016identity}, the identity mapping will be replaced by convolution layer when the resolution of feature maps is reduced, or when the dimension of feature channels are increased. 
This allows the networks to reset the variance of response to a small value, and avoid responses with very large variance, as shown in Figure.~\ref{fig:variance_accumulation}. 
The effect of increasing variance becomes more obvious in hourglass-like structures, where the responses of two residual units are summed together, as illustrated in Figure.~\ref{fig:Useconv}(a). 
Assume branches are uncorrelated, then the variance will be increased as:
{\small
\begin{align}
\Var\left[\bx^{(l+1)}\right] &= \sum_{i=1}^{2} \left( \Var\left[\bx^{(l)}_i\right] + \Var\left[\cF_i\left(\bx^{(l)}_i; \W^{(l)}_i\right)\right]  \right) \nonumber \\
&> \sum_{i=1}^{2} \Var\left[\bx^{(l)}_i\right].
\end{align}
}
\!\!Hence the output variance is almost doubled. When the network goes deeper, the variance will increase drastically.

In this paper, we use a $1\times 1$ convolution preceding with batch normalization and ReLU to replace the identity mapping when the output of two residual units are summed up, as illustrated in Figure.~\ref{fig:Useconv}(b).  
This simple replacement stops the variance explosion, as demonstrated in Figure.~\ref{fig:Useconv}(c). 
In experiments, we find that breaking the variance explosion also provide a better performance (Section~\ref{sec:ablation}).

\section{Experiments}

\subsection{Experiments on Human Pose Estimation}
\begin{table}
	\begin{footnotesize}
		\centering
		\caption{ Comparisons of PCKh@0.5 score on the MPII test set. 
			\textit{Ours-A} is trained using the training set used in~\cite{tompson2015efficient}. 
			\textit{Ours-B} is trained with the same settings but using all the MPII training set.  
			}
		\begin{tabular} 
			{@{}p{2.6cm}p{0.3cm}p{0.3cm}p{0.3cm}p{0.3cm}p{0.3cm}p{0.3cm}p{0.3cm}p{0.5cm}}
			\hline
			Method & Head & Sho. & Elb. & Wri. & Hip & Knee & Ank. & Mean\\
			\hline
			Pishchulin \etal~\cite{pishchulin2013strong} & 74.3  & 49.0  & 40.8  & 34.1  & 36.5  & 34.4 & 35.2 & 44.1  \\
			Tompson \etal~\cite{tompson2014joint}& 95.8  & 90.3  & 80.5  & 74.3  & 77.6  & 69.7 & 62.8 & 79.6  \\
			Carreira \etal~\cite{carreira2016human} & 95.7  & 91.7  & 81.7  & 72.4  & 82.8  & 73.2 & 66.4 & 81.3  \\
			Tompson \etal~\cite{tompson2015efficient}& 96.1  & 91.9  & 83.9  & 77.8  & 80.9  & 72.3 & 64.8 & 82.0  \\
			Hu\&Ramanan~\cite{hu2016bottom} & 95.0  & 91.6  & 83.0  & 76.6  & 81.9  & 74.5 & 69.5 & 82.4  \\
			Pishchulin \etal~\cite{pishchulin2016deepcut} & 94.1  & 90.2  & 83.4  & 77.3  & 82.6  & 75.7 & 68.6 & 82.4   \\
			Lifshitz \etal~\cite{lifshitz2016human} & 97.8  & 93.3  & 85.7  & 80.4  & 85.3  & 76.6 & 70.2 & 85.0   \\
			Gkioxary \etal~\cite{gkioxari2016chained} & 96.2  & 93.1  & 86.7  & 82.1  & 85.2  & 81.4 & 74.1 & 86.1   \\
			Rafi \etal~\cite{rafi2016efficient} & 97.2  & 93.9  & 86.4  & 81.3  & 86.8  & 80.6 & 73.4 & 86.3   \\
			Insafutdinov \etal~\cite{insafutdinov2016deepercut} & 96.8  & 95.2  & 89.3  & 84.4  & 88.4  & 83.4 & 78.0 & 88.5   \\
			Wei \etal~\cite{wei2016convolutional} & 97.8  & 95.0  & 88.7  & 84.0  & 88.4  & 82.8 & 79.4 & 88.5   \\
			Bulat\&Tzimiropoulos~\cite{bulat2016human} & 97.9  & 95.1  & 89.9  & 85.3  & 89.4  & 85.7 & 81.7 & 89.7   \\
			Newell \etal~\cite{newell2016stacked} & 98.2  & 96.3  & 91.2  & 87.1  & 90.1  & 87.4 & 83.6 & 90.9   \\
			\hline  
			Ours-A& 98.4  & 96.5  & 91.9  & 88.2  & 91.1  & 88.6 & 85.3 & 91.8  \\
			Ours-B& \textbf{98.5}  &\textbf{96.7}  & \textbf{92.5}  & \textbf{88.7}  & \textbf{91.1}  & \textbf{88.6} &\textbf{86.0} & \textbf{92.0}  \\
			\hline
		\end{tabular}
		\label{tab:MPII}
	\end{footnotesize}
\end{table}

\begin{table} 
	\begin{footnotesize}
		\centering
		\caption{   Comparisons of PCK@0.2 score on the LSP dataset.}
		\begin{tabular}{@{}p{2.7cm}p{0.3cm}p{0.3cm}p{0.3cm}p{0.3cm}p{0.3cm}p{0.3cm}p{0.3cm}p{0.4cm}}
			\hline
			Method & Head & Sho. & Elb. & Wri. & Hip & Knee & Ank. & Mean \\
			\hline 
			Belagiannis\&Zisserman~\cite{belagiannis2016recurrent} & 95.2 & 89.0 & 81.5 & 77.0 & 83.7 & 87.0 & 82.8 & 85.2 \\
			Lifshitz \etal~\cite{lifshitz2016human} & 96.8 & 89.0 & 82.7 & 79.1 & 90.9 & 86.0 & 82.5 & 86.7 \\
			Pishchulin \etal~\cite{pishchulin2016deepcut} & 97.0 & 91.0 & 83.8 & 78.1 & 91.0 & 86.7 & 82.0 & 87.1 \\
			Insafutdinov \etal~\cite{insafutdinov2016deepercut} & 97.4 & 92.7 & 87.5 & 84.4 & 91.5 & 89.9 & 87.2 & 90.1 \\
			Wei \etal~\cite{wei2016convolutional} & 97.8 & 92.5 & 87.0 & 83.9 & 91.5 & 90.8 & 89.9 & 90.5 \\
			Bulat\&Tzimiropoulos~\cite{bulat2016human} & 97.2 & 92.1 & 88.1 & 85.2 & 92.2 & 91.4 & 88.7 & 90.7 \\
			\hline
			Ours & \textbf{98.3}& \textbf{94.5} & \textbf{92.2} & \textbf{88.9} & \textbf{94.4} & \textbf{95.0} & \textbf{93.7} & \textbf{93.9} \\
			\hline      
		\end{tabular}
		\label{tab:LSP}
	\end{footnotesize}
\vspace{-1em}
\end{table}
We conduct experiments on two widely used human pose estimation benchmarks.
(i) The MPII human pose dataset~\cite{andriluka20142d}, which covers a wide range of human activities with 25k images containing over 40k people. 
(ii) The Leeds Sports Poses (LSP)~\cite{Johnson10} and its extended training dataset, which contains 12k images with challenging poses in sports. 

\subsubsection{Implementation Details}
Our implementation follows~\cite{newell2016stacked}. The input image is $256\times 256$ cropped from a resized image according to the annotated body position and scale. 
For the LSP test set, we simply use the image center as the body position, and estimate the body scale by the image size. 
Training data are augmented by scaling, rotation, flipping, and adding color noise. 
All the models are trained using Torch~\cite{collobert2011torch7}. 
We use RMSProp~\cite{tieleman2012lecture} to optimize the network on 4 Titan X GPUs with a mini-batch size of 16 (4 per GPU) for 200 epochs. The learning rate is initialized as $7\times10^{-4}$ and is dropped by 10 at the $150$th and the $170$th epoch. Testing is conducted on six-scale image pyramids with flipping. 

\subsubsection{Experimental Results}

\smalltitle{Evaluation measure.} Following previous work, we use the Percentage Correct Keypoints (PCK) measure~\cite{yang2013articulated} on the LSP dataset, and use the modified PCK measure that uses the matching threshold as $50\%$ of the head segment length (PCKh)~\cite{andriluka20142d} on the MPII dataset.

\begin{figure}[t]
\begin{center}
   \includegraphics[width=1\linewidth]{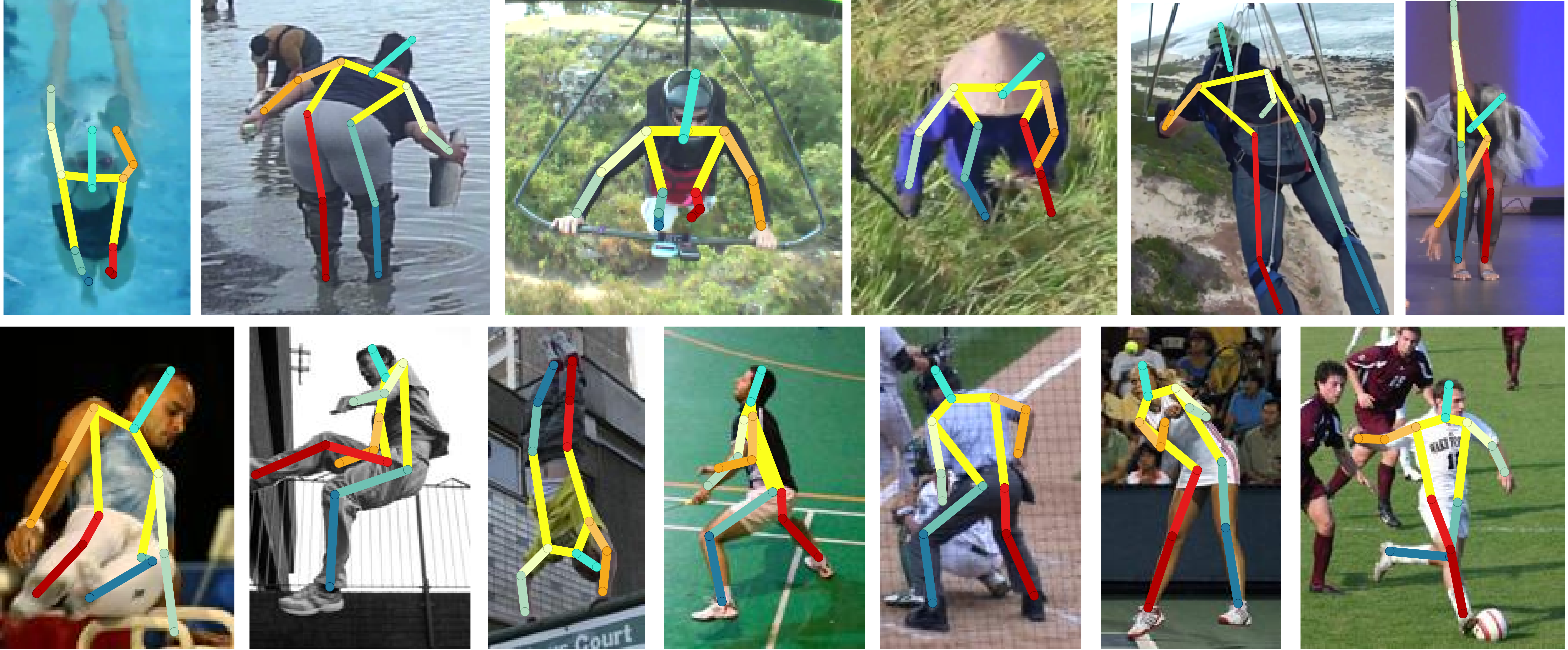}
\end{center}
	\vspace{-1em}
   \caption{ Results on the MPII (top) and the LSP dataset (bottom).}
	\label{fig:qualitative_results}
	\vspace{-1em}
\end{figure}

\smalltitle{MPII Human Pose.} 
We report the performance on MPII dataset in Table~\ref{tab:MPII}. 
Ours-A is trained using the training and validation set used in~\cite{tompson2015efficient}. 
Ours-B is trained with the same settings but using all the MPII training set. 
Our approach achieves $92.0\%$ PCKh score at threshold of $0.5$, which is the new state-of-the-art result. 
Specifically, our method achieves $1.6\%$ and $2.4\%$ improvements on \textit{wrist} and \textit{ankle}, which are considered as the most challenging parts to be detected. 
Qualitative results are demonstrated in Figure.~\ref{fig:qualitative_results}.

\textit{Complexity}. 
Our model increases the number of parameters by $13.5\%$ from  $23.7$M  to $26.9$M  given an eight-stack hourglass network. 
Our model needs $45.9$ GFLOPs for a $256\times256$ RGB image, which is a $11.4\%$ increase compared to hourglass network ($41.2$ GFLOPs). 
As reported in~\cite{newell2016stacked}, deeper hourglass with more stacks hardly improves result.

\smalltitle{LSP dataset.}
Table~\ref{tab:LSP} presents the PCK scores at the threshold of $0.2$. We follow previous methods~\cite{pishchulin2016deepcut,wei2016convolutional,insafutdinov2016deepercut} to train our model by adding MPII training set to the LSP and its extended training set. 
Our method improves the previous best result with a large margin by $3.2\%$. 
For difficult body parts, \eg, \textit{wrist} and \textit{ankle}, we have $3.7\%$ and $5.0\%$ improvements, respectively. 
Our method gains a lot due to the high occurrence of foreshortening and extreme poses presented in this dataset, as demonstrated in Figure.~\ref{fig:qualitative_results}. 

\subsubsection{Ablation Study}\label{sec:ablation}
\begin{figure}[t]
\begin{center}
   \includegraphics[width=1\linewidth]{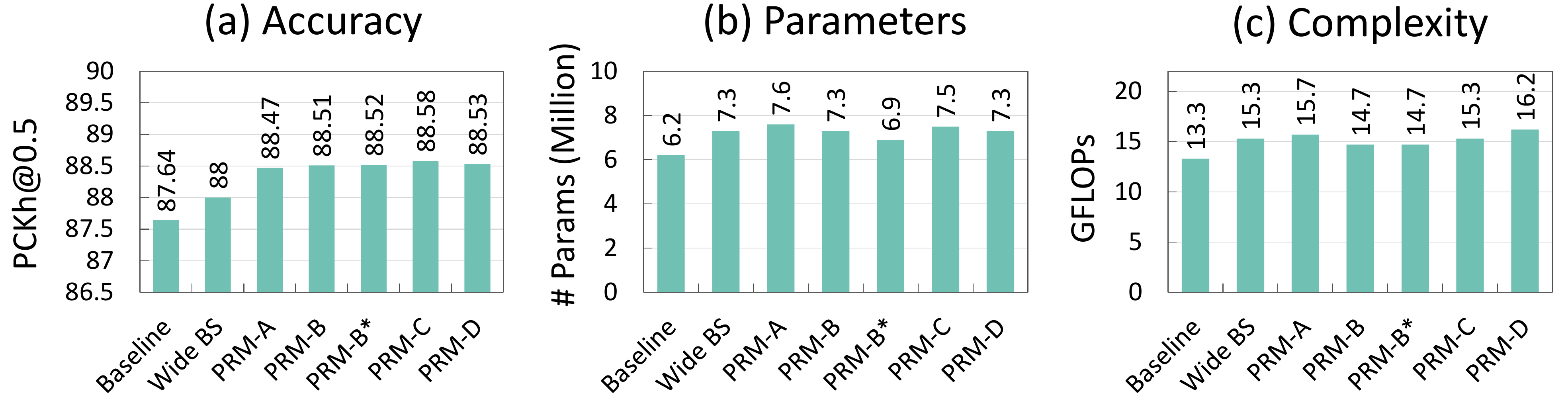}
\end{center}
   \vspace{-1em}
   \caption{Statistics of (a) accuracy, (b) number of parameters, and (c) computational complexity in terms of GFLOPs on different designs of PRMs in Figure.~\ref{fig:pyramidmodule}.}
\label{fig:ablation_arch}
\vspace{-1em}
\end{figure}

We conduct ablation study on the MPII validation set used in~\cite{tompson2015efficient} with a 2-stack hourglass network as the basic model. 

\smalltitle{Architectures of PRM.} 
We first evaluate different designs of PRM, as discussed in Section~\ref{sec:prm}, with the same number of branches, and the same feature channels for each branch (\eg, 5 branches with 28 feature channels for each pyramidal branch). 
We use PRM-A to PRM-D, which corresponds to Figure.~\ref{fig:pyramidmodule}, to denote the different architectures. 
Specifically, PRM-A produces separate input feature maps for different levels of pyramids, while PRM-B uses shared feature maps for all levels of pyramids. 
PRM-C uses concatenation instead of addition to combine features generated from pyramid, which is similar to inception models. 
PRM-D uses dilated convolutions, which are also used in ASPP-net~\cite{chen2016deeplab}, instead of pooling to build the pyramid. 
The validation accuracy is reported in Figure.~\ref{fig:ablation_arch}(a). 
All the PRMs have better accuracy compared with the baseline model. 
We observe that the difference in accuracy between PRM-A to PRM-D is subtle, while the parameters of PRM-A/C are higher than PRM-B/B*/D (Figure.~\ref{fig:ablation_arch}(b)), and the computational complexity (GFLOPs) of PRM-A/C/D are higher than PRM-B/B*. 
Therefore, we use PRM-B* in the rest of the experiments. 
Noted that increasing the number of channels to make the baseline model has the similar model size as ours (\textit{Wide BS})  would slightly improve the performance. But it is still worse than ours.

\begin{figure}
	\begin{center}
		\includegraphics[width=1\linewidth]{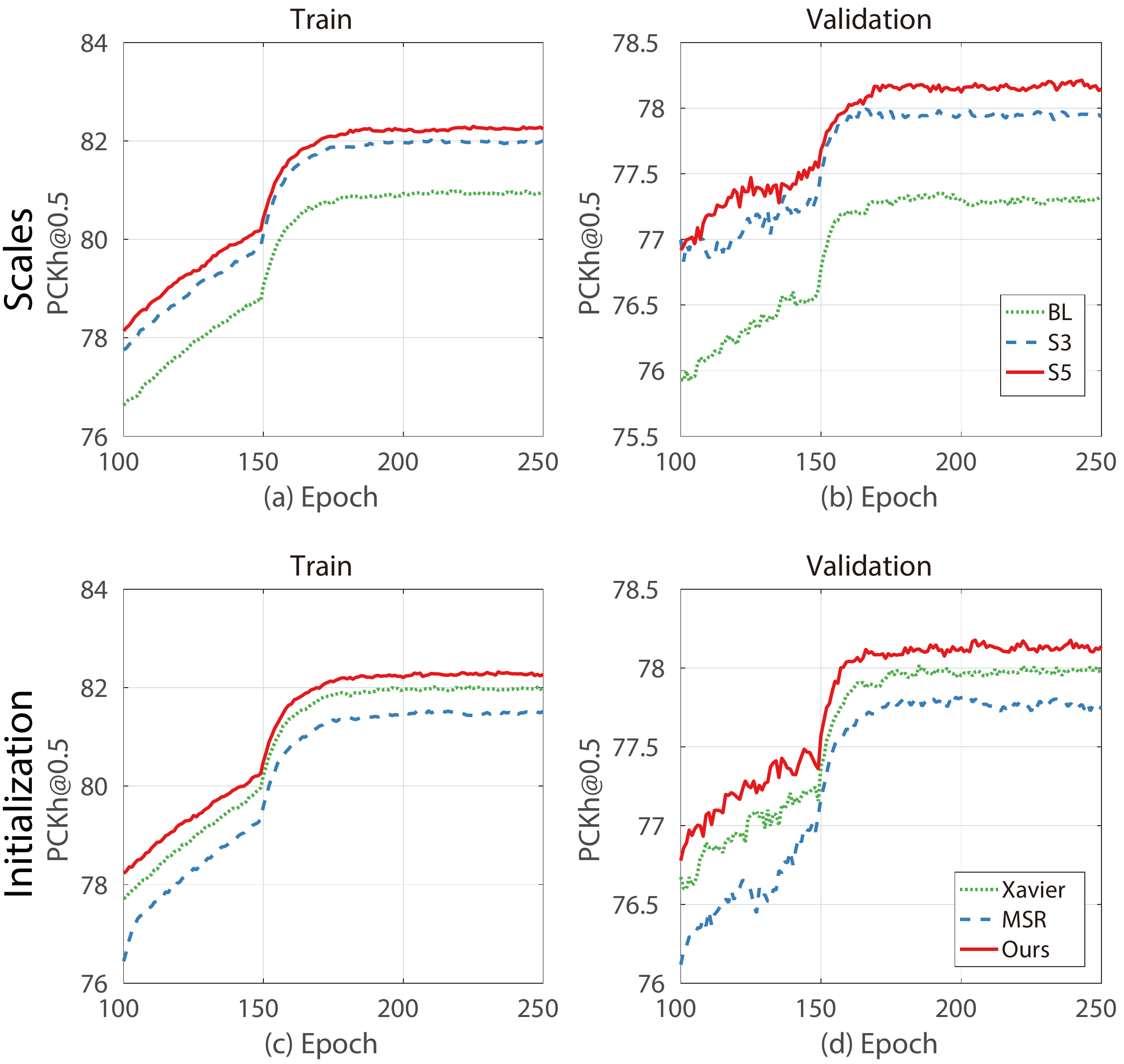}
	\end{center}
	\vspace{-1em}
	\caption{ Training and validation curves of PCKh scores \vs epoch on the MPII validation set. (a-b) Investigate the number of scales in the pyramid. BL stands for baseline model (two-stack hourglass network), S2 to S8 indicate PRM-B* with four scales to eight scales. 
	(c-d) Comparison of our initialization scheme with Xavier method~\cite{glorot2010understanding} and MSR method~\cite{he2015delving}. }
	\label{fig:scales_and_initialization}
	\vspace{-1.2em}
\end{figure}

\smalltitle{Scales of pyramids}. 
To evaluate the trade-off between the scales of pyramids $C$, we vary the scales from 3 to 5, and fix the model size by tuning the feature channels in each scale.  
We observe that increasing scales generally improves the performance, as shown in Figure.~\ref{fig:scales_and_initialization}(a-b). 

\smalltitle{Weight initialization.} 
We compare the performance of our initialization scheme with Xavier~\cite{glorot2010understanding} and MSR~\cite{he2015delving} methods. 
The training and validation curves of accuracy \vs epoch are reported in Figure~\ref{fig:scales_and_initialization}(c-d). 
It can be seen that the proposed initialization scheme achieves better performance than both methods. 



\smalltitle{Controlling variance explosion.} 
Controlling variance explosion, as discussed in Section~\ref{sec:control_va}, obtains higher validation score (88.0) compared with the baseline model (87.6). With our pyramid residual module, the performance could be further improved to 88.5 PCKh score.


\subsection{Experiments on CIFAR-10 Image Classification}

\begin{table}[]
\begin{small}
\centering
\caption{Top-1 test error ($\%$), model size (million) and GFLOPs on CIFAR-10. \textit{WRN-28-10} denote the Wide ResNet with depth 29 and widen factor 10. \textit{ResNeXt-29, $m\times n$d} denote ResNeXt with depth 29, cardinality $m$ and base width $n$. 
}
\label{table:cifar-comparison}
\begin{tabular} {lccc}
\hline
method      		& \#params	& GFLOPs 	& top-1 \\
\hline
WRN-28-10~\cite{zagoruyko2016WRN}         	
					& 36.5 		& 10.5		& 4.17         \\   
Ours-28-9        	& 36.4		& 9.5		& 3.82	   	   \\ 
Ours-28-10    		& 42.3 		& 11.3 		& 3.67         \\
\hline  
ResNeXt-29, $8\times 64$d~\cite{xie2016aggregated}
					& 34.4  	& 48.8 		& 3.65        \\   
ResNeXt-29, $16\times 64$d~\cite{xie2016aggregated}  
					& 68.2  	& 184.5 	& 3.58        \\   
Ours-29, $8\times 64$d 		
					& 45.6  	& 50.5		& 3.39        \\ 
Ours-29, $16\times 64$d		
					& 79.3	  	& 186.1		& \textbf{3.30}    	        \\ 
\hline 
\end{tabular}
\end{small}
\vspace{-0.2em}
\end{table}

The CIFAR-10 dataset~\cite{krizhevsky2009learning} consists of 50k training images and 10k test images with size $32 \times 32$ drawn from 10 classes. 
We follow previous works for data preparation and augmentation. 
We incorporate the proposed pyramid branches into two state-of-the-art network architectures, \ie, Wide residual networks~\cite{zagoruyko2016WRN} and ResNeXt~\cite{xie2016aggregated}. 
We add four pyramid branches with scales ranging from $0.5$ to $1$ into the building block of both Wide ResNet and ResNeXt. 
For Wide ResNet, the total width of all pyramid branches is equal to the width of the output of each residual module.   
For ResNeXt, we simply use the same width as its original branches for our pyramid branches. 
Table~\ref{table:cifar-comparison} shows the top-1 test error, model sizes and GFLOPs. 
Our method with similar or less model size (\textit{Ours-28-9 \vs WRN-28-10} and \textit{Ours-29, $8\times 64$d \vs ResNeXt-29, $16\times 64$d}) achieve better results. A larger model with our pyramid module (\textit{Ours-29, $16\times 64$d	}) achieves $3.30\%$ test error, which is the state-of-the-art result on CIFAR-10. 

\section{Conclusion}
This paper has proposed a Pyramid Residual Module to enhance the invariance in scales of the DCNNs. 
We also provide a derivation of the initialization scheme for multi-branch networks, and demonstrate its theoretical soundness and efficiency through experimental analysis. 
Additionally, a simple yet effective method to prevent the variances of response from explosion when adding outputs of multiple identity mappings has been proposed. 
Our PRMs and the initialization scheme for multi-branch networks are general, and would help other tasks.

\smalltitle{Acknowledgment}: 
This work is supported by SenseTime Group Limited, the General Research Fund sponsored by the Research Grants Council of Hong Kong (Project Nos. CUHK14213616, CUHK14206114, CUHK14205615, CUHK419412, CUHK14203015, CUHK14207814, and CUHK14239816), the Hong Kong Innovation and Technology Support Programme (No.ITS/121/15FX), National Natural Science Foundation of China (No. 61371192), and ONR N00014-15-1-2356. 

{\small
\bibliographystyle{ieee}
\bibliography{pose}
}

\end{document}